%% file: impact-of-depth.tex
\pdfoutput=1
\documentclass[11pt]{article}

\usepackage{acl2023}
\usepackage{times}
\usepackage{newtxmath}
\usepackage[T1]{fontenc}
\usepackage[utf8]{inputenc}
\usepackage{microtype}
\usepackage{inconsolata}

\input{preamble}

\title{The Impact of Depth on Compositional Generalization in \linebreak[1] Transformer Language Models}
\author{Jackson Petty$^{\dagger}$\thanks{~ Work completed as part of a Google Student Researcher project.}\quad Sjoerd van Steenkiste$^{\S}$\quad Ishita Dasgupta$^{\star}$\quad Fei Sha$^{\S}$ \\ \bfseries Dan Garrette$^{\star}$\quad Tal Linzen$^{\S}$  \\ \\
${}^{\dagger}$ New York University\quad ${}^{\S}$ Google Research\quad ${}^{\star}$ Google DeepMind\\
\texttt{petty@nyu.edu}\quad \texttt{\{svansteenkiste,idg,fsha,dhgarrette,linzen\}@google.com}
}

\begin{document}

\input{paper}

\end{document}

%% file: paper.tex
\maketitle

\begin{abstract}
To process novel sentences, language models (LMs) must generalize compositionally---combine familiar elements in new ways. What aspects of a model's structure promote compositional generalization? Focusing on transformers, we test the hypothesis, motivated by theoretical and empirical work, that deeper transformers generalize more compositionally. Simply adding layers increases the total number of parameters; to address this confound between depth and size, we construct three classes of models which trade off depth for width such that the total number of parameters is kept constant (\SI{41}{\M}, \SI{134}{\M} and \SI{374}{\M} parameters). We pretrain all models as LMs and fine-tune them on tasks that test for compositional generalization. We report three main conclusions: (1) after fine-tuning, deeper models generalize more compositionally than shallower models do, but the benefit of additional layers diminishes rapidly; (2) within each family, deeper models show better language modeling performance, but returns are similarly diminishing; (3) the benefits of depth for compositional generalization cannot be attributed solely to better performance on language modeling. Because model latency is approximately linear in the number of layers, these results lead us to the recommendation that, with a given total parameter budget, transformers can be made shallower than is typical without sacrificing performance.
\end{abstract}

\section{Introduction}

The number of possible sentences in natural language is enormous; regardless of the size of its training set, a language model (LM) will regularly encounter sentences it has never seen before. The ability to interpret such sentences relies on compositional generalization: the capacity to combine familiar words and syntactic structures in new ways \citep{montague-1970-UniversalGrammar,fodor-1988-ConnectionismAnalysis}. Transformer LMs \citep{vaswani-2017-AttentionNeed}, while highly successful in many settings, often struggle when tested on benchmarks that require compositional generalization \citep{kim-2020-COGSInterpretation}. What architectural factors affect a transformer's ability to generalize compositionally?

In this paper, we test the hypothesis that increasing a transformer's depth---the number of layers it has---improves its performance on tasks that require compositional generalization. This hypothesis is motivated both by theoretical work, which has shown that adding layers increases the expressive capacity of neural networks in general \citep{raghu-2017-ExpressiveNetworks} and transformers in particular \citep{merrill-2021-SaturatedCircuits}, and by experimental work suggesting that deeper models generalize more compositionally than shallower ones \citep{mueller-2022-ColoringModels,murty-2022-CharacterizingProjections}. 

While existing empirical work lends some credibility to this hypothesis, to directly confirm it we must address the confound between depth and size (number of parameters). As each additional layer introduces a new set of parameters, deeper models are also larger, all else being equal, and LMs' performance on a wide variety of tasks is known to be correlated with their size \citep{kaplan-2020-ScalingModels,hoffmann-2022-TrainingModels,muennighoff-2023-ScalingModels}. To address this confound, we construct classes of models with an equal total number of parameters but differing depths; we do so by reducing the model's feed-forward dimension to compensate for added depth. We pretrain all models as language models and fine-tune them on four compositional generalization tasks: the semantic parsing tasks COGS \citep{kim-2020-COGSInterpretation}, COGS-vf \citep{qiu-2022-ImprovingAugmentation} and GeoQuery \citep{zelle-1996-LearningProgramming}, and the English passivization portion of Multilingual Transformations \citep{mueller-2022-ColoringModels}. In all of these tasks, the model is trained on a particular data distribution and is expected to generalize to another distribution by composing together familiar elements in novel ways.

In addition to any possible direct effect on compositional generalization, depth may also be correlated with other factors which may themselves predict compositional generalization, such as language modeling loss during pretraining or in-distribution performance on the fine-tuning task. This complicates the interpretation of any relationship we might find between depth and generalization performance. To address this concern, we also investigate and correct for the effect of depth on language modeling performance and in-distribution loss.

We report the following findings, across three model size classes (\SI{41}{\M}, \SI{134}{\M}, and \SI{374}{\M} parameters):
\begin{enumerate}[leftmargin=*]
\item In general, deeper models have lower perplexity (\Cref{sec:pretraining}). The marginal increase in performance gained from additional layers diminishes rapidly as models get deeper, and performance begins to degrade when the feed-forward dimension approaches the dimensionality of the model's contextualized embeddings.
\item In general, deeper models display better compositional generalization (\Cref{sec:compositionality}). Again, most of the benefit of depth accrues from the first few layers; for several of the compositional generalization benchmarks we use, performance saturates very quickly as models get deeper.
\item Deeper models generalize more compositionally even after correcting for the fact that their language modeling perplexity is lower and their in-distribution performance on the fine-tuning task is higher (\Cref{sec:id-correction}). 
\item Since transformers' latency is approximately linear in their depth, it is in many case more efficient to make a model wider rather than deeper, given a fixed parameter budget  (\Cref{sec:compute-cost}).
\end{enumerate}

\section{Methodology} \label{sec:methods}

\subsection{Constructing Families of Models with Equal Numbers of Parameters}

To make a transformer deeper without increasing the total number of parameters, we need to also make it narrower. There are several ways to do so: we can reduce the size of the feed-forward dimension $\dff$, reduce the size of the contextual embeddings $\dmodel$, or reduce the size of the attention outputs $\dattn$ (see \Cref{sec:layer-diagram} for a diagram of a transformer layer annotated with dimensionality labels).
\citet{vaswani-2017-AttentionNeed} coupled these variables at $\dmodel = \dattn = \dff / 4$. Most transformer LMs have adopted this ratio \citep[\emph{inter alia}]{devlin-2019-BERTUnderstanding,kaplan-2020-ScalingModels,hoffmann-2022-TrainingModels}, though \citet{raffel-2019-ExploringTransformer} increased the size of $\dff$ relative to $\dmodel$ and $\dattn$ for their two largest models. By contrast, we vary $\dff$ with depth (while holding $\dmodel = \dattn$ constant). By keeping the attention mechanism identical across models of varying depths, we rule out the possibility that depth will be confounded with the capacity of the self-attention mechanism. We refer to $\dmodel/\dff$, conventionally set to $1/4$, as the \emph{feed-forward ratio}. 

\begin{figure*}[!ht]
    \floatbox[{\capbeside\thisfloatsetup{capbesideposition={right,top},capbesidewidth=6cm}}]{figure}[\FBwidth]
    {\vspace{-4pt}\caption{Models for the 41M-, 134M-, and 374M-parameter size classes. Points indicate models trained in this paper, and black diamonds represent the baseline models for each class whose hyperparameters were taken from \citet{kim-2020-COGSInterpretation} and \citet{raffel-2019-ExploringTransformer}.}\label{fig:iso_curve}}
    {\includegraphics[height=1.4in]{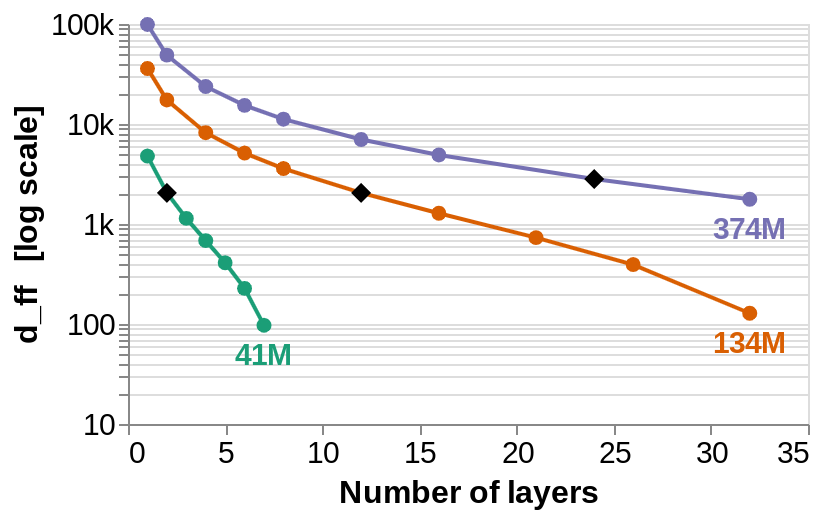}}
\end{figure*}

\paragraph{Deriving hyperparameter relations.} As a starting point for our size classes of models, we use hyperparameters taken from T5-base and T5-large \citep{raffel-2019-ExploringTransformer} as well as a smaller model from \citet{kim-2020-COGSInterpretation} which has identical layer-internal hyperparameters to T5-small but fewer layers.\footnote{Unlike T5 and the original transformer, we implement GPT-style causal decoder-only language models; following \citet{wang-2022-What} we consider decoder-only models with half as many total layers as their encoder-decoder variants.} We implement models using \texttt{t5x} \cite{roberts2022scaling}. We then calculate how much the feed-forward dimension must change to accommodate adding or removing layers.
Starting from the formula in \citet{kaplan-2020-ScalingModels}, the number of parameters $M$ in a layer is
\begin{equation*}
    M(\dff) = 2\dmodel\dff + 4\dmodel\dattn = \beta \cdot \dff + A,
\end{equation*}
where the constant $\beta$ represents the contribution of the parameters of the feed-forward block which projects vectors from $\mathbb{R}^{\dmodel}$ into $\mathbb{R}^{\dff}$ and back into $\mathbb{R}^{\dmodel}$;
and the constant $A$ represents the parameters of everything aside from the feed-forward block, including the attention mechanism.
The total parameter count of a full model $N$ in terms of $\dff$ and $\nlayers$ is then
\begin{align*}
    N(\nlayers,\dff) &= \nlayers \cdot M(\dff) + 2\dmodel\nvocab.
\end{align*}
Given initial values $(\nlayers^0, \dff^0)$ which characterize the baseline model in each size class (e.g., T5-large), our goal is to find pairs $k, w(k)$ such that
\begin{equation*}
    N(\nlayers^0 + k, \dff^0 - w(k)) = N(\nlayers^0,\dff^0).
\end{equation*}
Solving for $w$ as a function of $k$ tells us how much to increase (or decrease) $\dff^0$ if we remove (or add) $k$ layers from an existing model:
\begin{equation}
    w(k) = \left\lfloor\left(1-\frac{\nlayers^0}{\nlayers^0 + k}\right)\left(\dff^0 + \frac{A}{\beta}\right)\right\rceil. \label{eq:w-k}
\end{equation}
Since adding or removing $k$ layers might require changing $\dff^0$ by a fractional amount, we round $w(k)$ to the nearest integer.
\Cref{tab:iso_classes} reports the exact hyperparameter values we use for each of our three size classes, derived from \Eqref{eq:w-k} above, and \Cref{fig:iso_curve} shows each size class plotted as $(\nlayers,\dff)$ pairs.

\subsection{Datasets and Training}

\begin{table*}[!ht]
    \centering \small
    \begin{tabularx}{\textwidth}{lX}
        \toprule 
        COGS & $x\colon$ A hedgehog ate the cake . \\ & $y\colon$ ${}^*\textrm{cake}(x_4);\textrm{hedgehog}(x_1) \,\textsc{and}\,\textrm{eat.agent}(x_2, x_1)\,\textsc{and}\,\textrm{eat.theme}(x_2,x_4)$ \\
        \midrule 
        COGS-vf & $x\colon$ A hedgehog ate the cake on the bed . \\
                & $y\colon$ $\textrm{eat} ( \textrm{agent} = \textrm{hedgehog} , \textrm{theme} = {}^* \textrm{cake} ( \textrm{nmod} . \textrm{on} = {}^* \textrm{bed} ) )$ \\
        \midrule 
        GeoQuery & $x\colon$ which states have cities named m0 \\
                 & $y\colon$ $\textrm{answer} ( \textrm{intersection} ( \textrm{state} , \textrm{loc\_1} ( \textrm{intersection} ( \textrm{city} , \textrm{m0} ) ) ) )$ \\
        \midrule
        English Passivization & $x\colon$ our vultures admired her walrus above some zebra .\\
        & $y\colon$ her walrus above some zebra was admired by our vultures . \\ \bottomrule
    \end{tabularx}
    \caption{Examples of inputs ($x$) and targets ($y$) from each compositional generalization dataset.}
    \label{tab:datasets}
\end{table*}

\subsubsection{Language Modeling}

We use the Colossal Clean Crawled Corpus (C4; \citealt{raffel-2019-ExploringTransformer}) for pretraining. 
We use a context size $\nctx$ of \num{1024} tokens and a batch size of \num{128} sequences $\approx$ \SI{131}{\k} tokens. We pretrain each model for \SI{1}{\M} steps, resulting in a total training dataset of roughly \SI{131}{\B} tokens.

\subsubsection{Compositional Generalization}

In compositional generalization datasets, models are tested on a distribution that contains novel combinations of pieces, each of which has been previously seen independently during training.
We fine-tune our pretrained models on the training portion of the dataset for \num{10000} steps with a batch size of 128.
Validation loss continued to decrease throughout training runs on each dataset, so we report values from the end of each fine-tuning run without early stopping.  We use the following datasets (for examples of instances of these tasks, see~\Cref{tab:datasets}): 

\begin{enumerate}[leftmargin=*]
    \item \textbf{COGS} \citep{kim-2020-COGSInterpretation} is a semantic parsing dataset consisting of natural-language sentences paired with formal semantic representations. It is constructed such that the out-of-domain generalization distribution contains two generalization types: new combinations of familiar words (\emph{lexical generalization}, such as using the word `hedgehog' as the object of a sentence when this word has only been seen during training as a subject); or new combinations of familiar syntactic structures (\emph{structural generalization}, such as relative clauses that are more deeply nested than seen in training).

    \item \textbf{Variable-free COGS} (COGS-vf; \citealt{qiu-2022-ImprovingAugmentation}) is a simplified variant of COGS where the semantic representations do not use numbered variables (see \Cref{tab:datasets} for a comparison between COGS and COGS-vf). Removing variables from the representation has the benefit of lowering the associated computational cost of training by making sequences shorter. This conversion has been previously shown to improve the performance of models by reducing the complexity of the output space \citep{qiu-2022-EvaluatingParsing}, but comes at the cost of limiting the capacity of the formal language to represent phenomena that require coordination of variable identity, such as control and anaphor binding.

\item \textbf{GeoQuery} \citep{zelle-1996-LearningProgramming} contains natural-language questions about US geography paired with SQL-style database queries representing those questions. We report results on the GeoQuery Standard split.

\item \textbf{English passivization} \citep{mueller-2022-ColoringModels} is a dataset of English active-voice sentences paired with their passive-voice counterparts (adapted from \citealt{mulligan-2021-StructureTransformations-a}). This benchmark is designed to test whether models use shallow, positional heuristics or syntactically principled ones. While \citet{mueller-2022-ColoringModels} implemented a number of transformations in different languages, we focus on the English Passivization task. 

\end{enumerate}

\section{Results}

\subsection{Language Modeling} \label{sec:pretraining}

\begin{figure*}[t]
    \centering
    \includegraphics[height=1.4in]{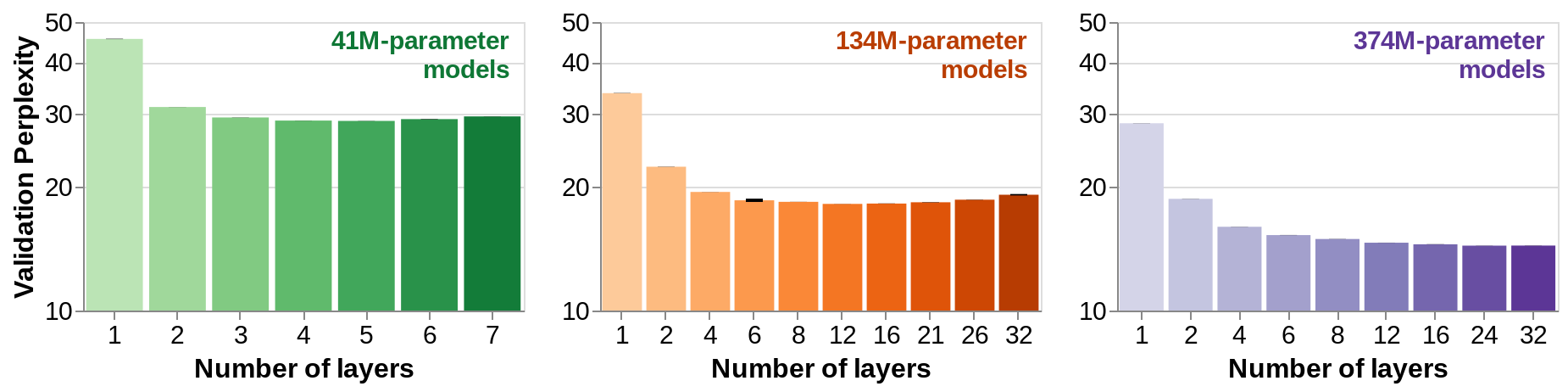}
    \caption{Deeper models achieve lower perplexities than shallower ones after equal amounts of training data regardless of size, but the benefits of adding layers diminish quickly with depth. Mean over 5 runs shown with error bars.}
    \label{fig:pretrain_bar}
\end{figure*}

\begin{figure*}
    \centering
    \begin{subfigure}[b]{0.45\textwidth}
        \centering
        \includegraphics[height=1.4in]{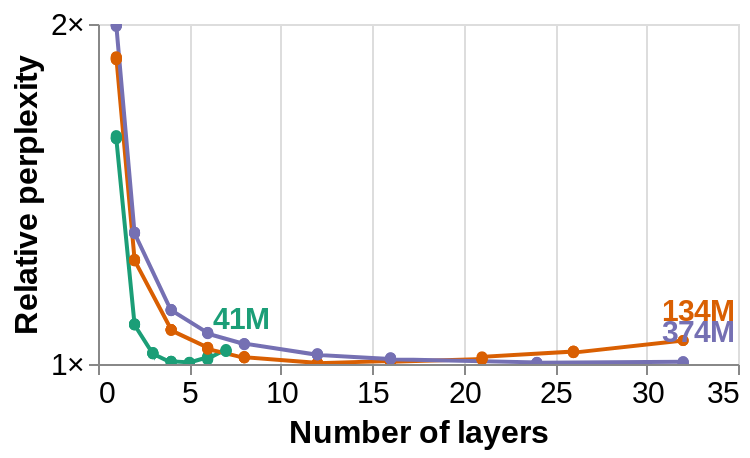}
        \caption{}
        \label{fig:ppl_by_depth}
    \end{subfigure}
    \hspace{-2.5em}
    \begin{subfigure}[b]{0.45\textwidth}
        \centering
        \includegraphics[height=1.4in]{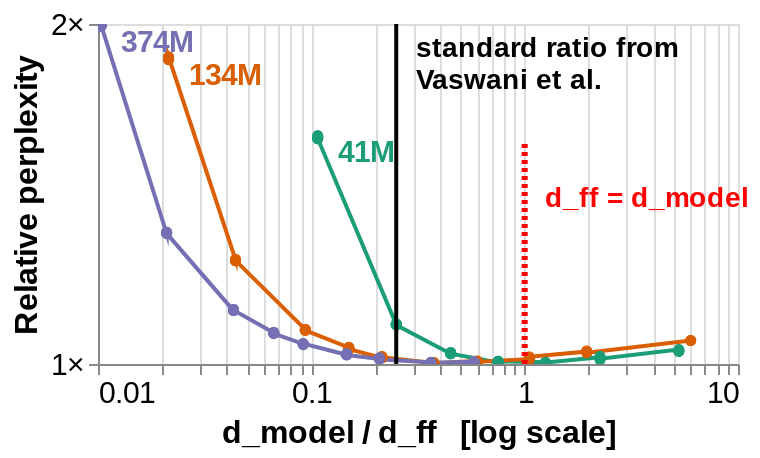}
        \caption{}
        \label{fig:ppl_by_ratio}
     \end{subfigure}
    \caption{Relative perplexity compared to the best model in each size class. \textbf{(a)} Perplexity goes down rapidly as models get deeper; only a few layers are needed to obtain most of the value of depth. \textbf{(b)} When $\dmodel/\dff > 1$ (red dashed rule), perplexity slowly increases. As models get larger, the range of $\dmodel/\dff$ ratios where performance is close-to-optimal expands leftward to include smaller and smaller values.}
    \label{fig:relative_ppl}
\end{figure*}

\paragraph{Deeper models have lower perplexity.} Depth has a significant impact on language modeling performance. At the shallow end of the spectrum, increasing depth results in a dramatic improvement in perplexity (\Cref{fig:pretrain_bar}). In \Cref{fig:ppl_by_depth} we compare the perplexity of each model in a size class relative to that of the best-performing model of that size. In the extreme case, the perplexity of a single-layer model can be nearly twice that of the optimal model in the class. Moreover, as parameter count increases the disparity between the worse, shallower models and the better, deeper models increases as well: For \SI{41}{\M}-parameter models the ratio between the perplexity of the single-layer model and that of the optimal (5-layer) model is \num{1.59}; for the \SI{134}{\M}-parameter models, the ratio is \num{1.86}; and for the \SI{374}{\M}-parameter models, the ratio is \num{1.99}.

\paragraph{Performance increases most rapidly within the first few layers.} While deeper models do, in general, perform better than shallower ones, the increase in performance that comes from adding layers diminishes rapidly as models become deeper (\Cref{fig:ppl_by_depth}). The performance difference between 1-layer and 2-layer models is dramatic across all size classes; moving from 2 to 4 layers results in a much more modest performance improvement. We also note that as models get larger in our setup, they are able to make productive use of increasingly more layers: the optimal \SI{41}{\M}-parameter model in our setup has 5 layers, while the optimal 134M-parameter model has 12; among \SI{374}{\M}-parameter models, the 24-layer model had the best performance. At the same time, the pattern of the diminishing utility of depth holds even for the largest models we study.

\paragraph{Performance degrades when models become too narrow.} At the deeper end of our scale, adding layers is not only unhelpful for performance, but begins to harm it (see the right-hand sides of each size-class curve in \Cref{fig:ppl_by_depth}). As previously noted, the point at which trading width for depth becomes harmful is not an absolute function of depth, since the depth of the optimal model differs across classes.
However, comparing the relative performance of models within a size class to the feed-forward ratio $\dmodel/\dff$ shows that model performance begins to worsen once $\dff$ becomes smaller than $\dmodel$ (to the right of the red dashed line in \Cref{fig:ppl_by_ratio}); when this happens, the affine projection of the vectors from $\mathbb{R}^{\dmodel}$ into $\mathbb{R}^{\dff}$ becomes a non-injective map. In \Cref{sec:rank} we analyze the weight matrices of the affine transforms in the feed-forward network of each layer and demonstrate that as $\dmodel/\dff$ increases the transforms become increasingly rank-deficient.

\paragraph{Larger models can tolerate more extreme feed-forward ratios.} Varying $\dff$ while keeping $\dmodel$ constant results in feed-forward ratios $\dmodel/\dff$ which deviate significantly from the ratio of $1/4$, which is the de-facto standard in the literature (black vertical rule in \Cref{fig:ppl_by_ratio}). We find that smaller models are more sensitive to the particular value of the feed-forward ratio, and that for small models the standard ratio may not be optimal. Within the \SI{41}{\M}-parameter size class there is a narrow range of feed-forward ratios in which model performance is within a few percentage points of the best-in-class model. As models get larger, this range expands leftward to include models which have increasingly wide feed-forward networks relative to the size of their contextual embeddings. In other words, larger models have more leeway to trade depth for width, becoming wider in proportion to their model dimension $\dmodel$ without incurring large penalties for their perplexity. Furthermore, when $\dmodel/\dff < 1$ the feed-forward ratio does not predict the relative perplexity of a model independent of its size.

\subsection{Compositional Generalization} \label{sec:compositionality}

We next fine-tune the models pretrained in the previous section on the training portions of each compositional generalization dataset, and measure the full-sequence (exact match) accuracy of the models on the out-of-distribution generalization set.

\begin{figure*}
    \centering
    \includegraphics[height=1.4in]{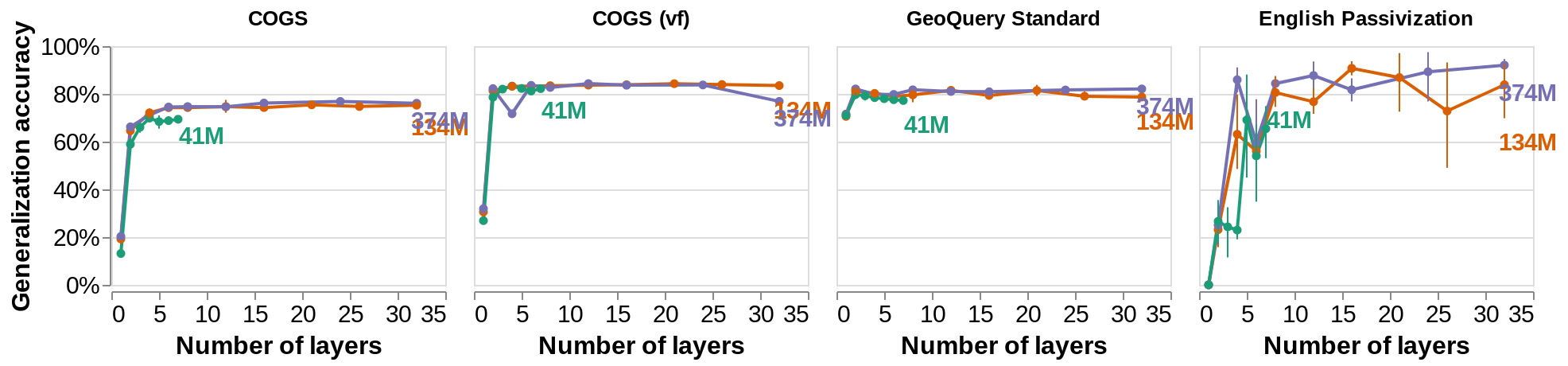}
    \caption{Deeper models generalize better than shallower models on compositional tasks across datasets and size classes. Error bars (easily visible only on the English Passivization data) report 95\% confidence intervals in estimation of the mean, taken over 5 runs.
    }
    \label{fig:comp_lines}
\end{figure*}

\paragraph{Deeper models generalize better.} As with language-modeling performance, deeper models tend to attain higher generalization accuracies than shallower models in the same size class (\Cref{fig:comp_lines}). The effect of depth on compositional generalization is more variable than it is for language modeling, however: for COGS, COGS-vf, and GeoQuery there is some small non-monotonicity in the generalization accuracy as a function of depth. On English Passivization, the \SI{41}{\M}- and \SI{134}{\M}-parameter classes show largely-consistent trends where deeper models perform better than shallower ones; the \SI{374}{\M}-parameter models show more significant non-monotonicity, though the deepest models do still outperform the shallowest ones.

\paragraph{The benefit of depth saturates quickly for some tasks.} As with language modeling, most of the benefit of depth is gained by having only a few layers. 
For three of the tasks---COGS, COGS-vf, and GeoQuery---we see threshold depths after which generalization accuracy stays relatively constant as depth increases. These threshold depths are low and constant across model sizes, but vary by dataset: 4--6 layers for COGS, and 2--4 layers for COGS-vf and GeoQuery. Performance on \mbox{COGS-vf} appears to saturate with fewer layers than on COGS despite the fact that the two datasets
express the same linguistic phenomena;
this suggests that the saturation we observe on some datasets is closely linked to the complexity of the output representation independent from the complexity of the compositional generalization expressed in the data. On English Passivization, the impact of depth is more variable, which makes it difficult to identify a size-independent threshold.

\begin{figure*}
    \centering
    \includegraphics[height=1.4in]{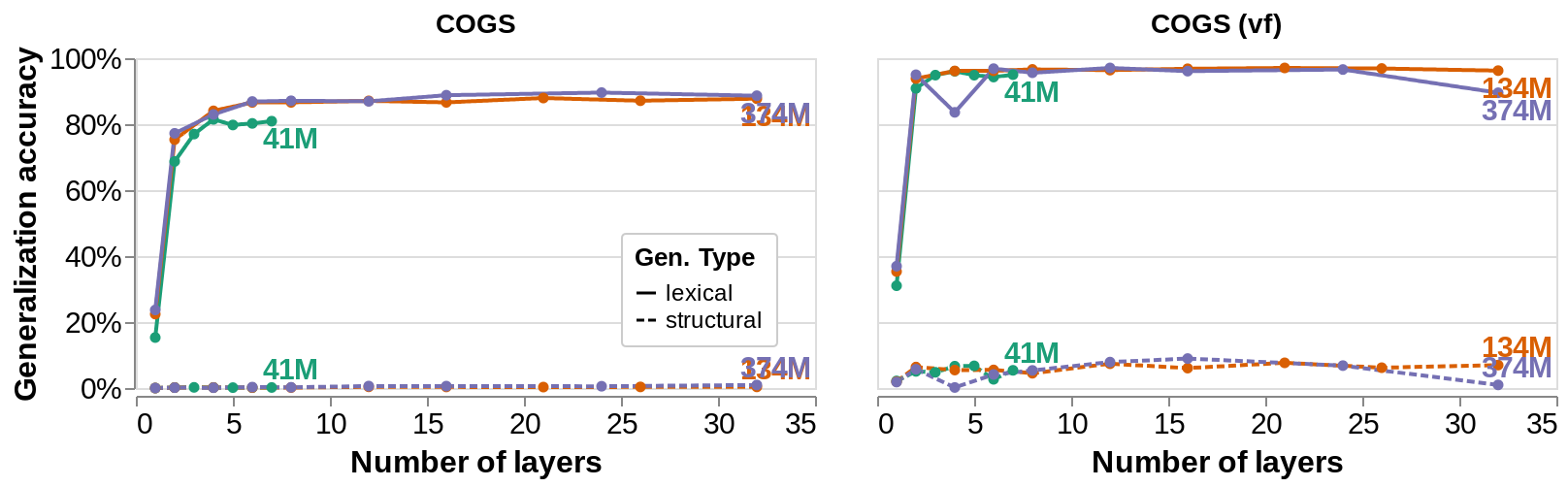}
    \caption{Increasing depth improves lexical generalization (solid lines) in both COGS and COGS-vf, but does not meaningfully improve structural generalization performance (dashed lines). Data shown is from a single run per condition.}
    \label{fig:struct_lex}
\end{figure*}

The threshold effects suggest that some subsets of the datasets can be addressed with relatively simple models. We investigate this hypothesis by separately analyzing the models' performance on the two types of generalization cases included in COGS and COGS-vf: lexical generalization, where a familiar word needs to be interpreted in a familiar syntactic context in which it has not been observed; and structural generalization, where the syntactic structure is novel and needs to be constructed from familiar syntactic pieces. We find that even deep models at the largest model size systematically fail to generalize structurally (\Cref{fig:struct_lex}); the benefit of depth is largely limited to the easier lexical generalization cases. This supports the hypothesis that the saturated effect of depth is due to the existence of easier subsets of the datasets, and shows that increasing depth alone does not substantially improve the models' ability to learn the correct inductive bias for these structural tasks. 

\subsection{Depth Effects are Independent between Upstream and Downstream Tasks} \label{sec:id-correction}

\begin{figure*}[t]
    \begin{subfigure}[b]{0.4\textwidth}
        \centering
        \includegraphics[height=1.5in]{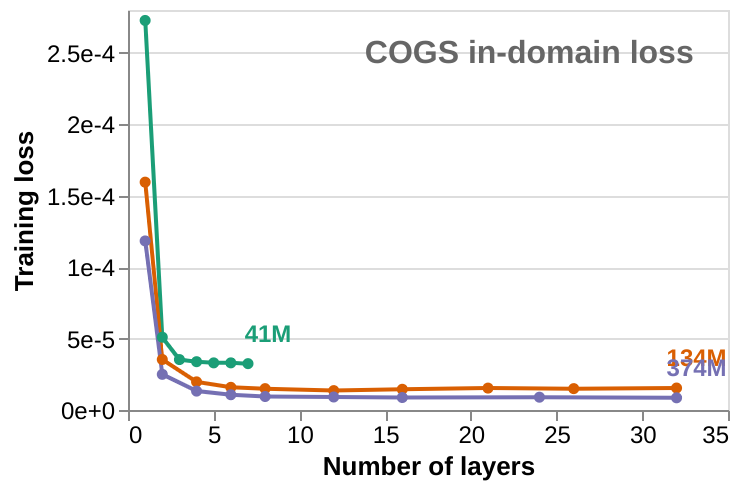}
        \caption{}
        \label{fig:cogs-id-loss}
    \end{subfigure}
    \quad
    \begin{subfigure}[b]{0.4\textwidth}
        \centering
        \includegraphics[height=1.5in]{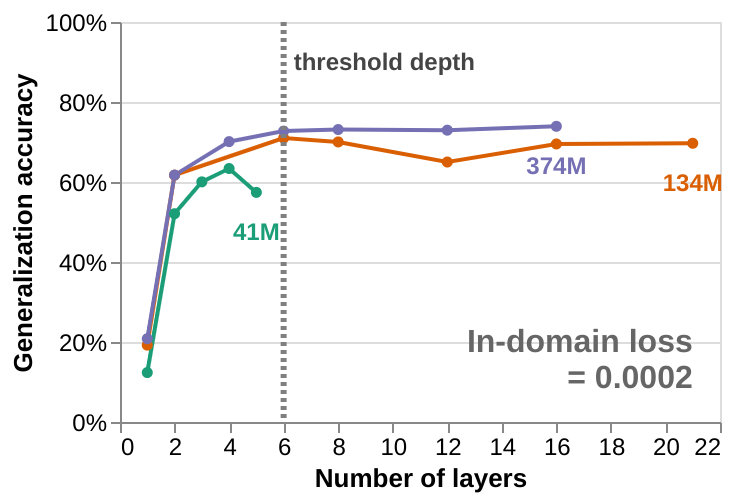}
        \caption{}
        \label{fig:loss_corr_ft}
    \end{subfigure}
    \caption{\textbf{(a)} Deeper models attain lower (better) in-domain loss values on compositional tasks. \textbf{(b)} Deeper models generalize better than shallower ones on COGS, even at points during fine-tuning when models have equal loss (0.0002) on the in-distribution portion of the dataset. Data shown is from a single run per condition.}
\end{figure*}

We have shown that deeper models generalize better than shallower models do. But in \Cref{sec:pretraining} we also showed that deeper models attain lower validation perplexity in pretraining than shallower models; and deeper models achieve lower in-distribution loss on the fine-tuning tasks than shallower ones (\Cref{fig:cogs-id-loss}). Both of these observations constitute potential confounds for the interpretation of the previous section: it could be that depth does not \emph{directly} improve generalization accuracy, but only does so indirectly by improving language modeling performance or in-distribution accuracy on the fine-tuning task, which in turn lead to better generalization. To determine if this is the case, we correct for both of these potential confounds. 

\begin{figure*}[t]
    \centering
    \begin{subfigure}[b]{0.4\textwidth}
        \centering
        \includegraphics[height=1.5in]{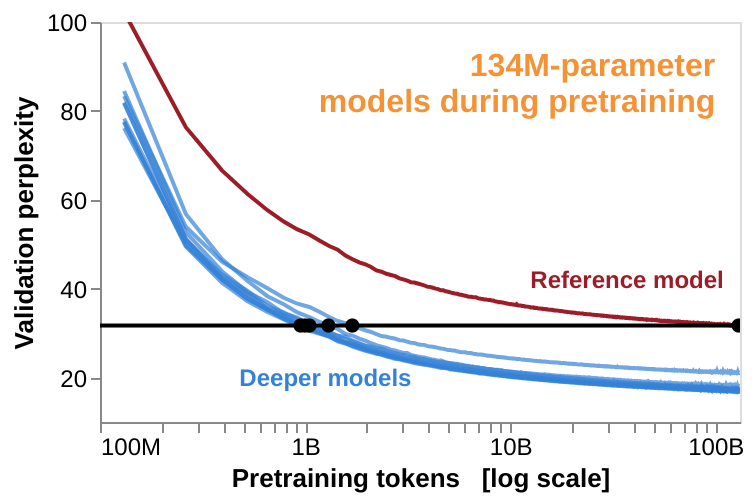}
        \caption{}
        \label{fig:equal_ppl_points}
    \end{subfigure}
    \quad
    \begin{subfigure}[b]{0.4\textwidth}
        \centering
        \includegraphics[height=1.5in]{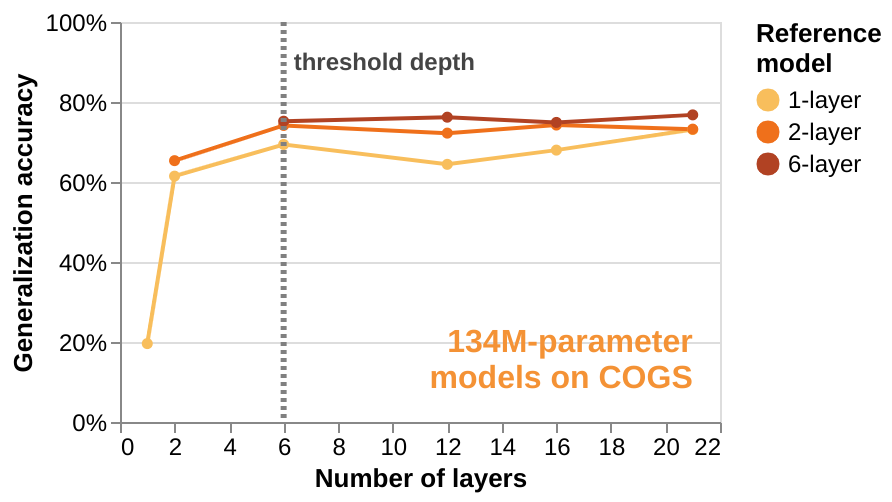}
        \caption{}
        \label{fig:loss_corr_pt}
    \end{subfigure}
    \caption{\textbf{(a)} To correct for the potential effect of deeper models' lower pretraining loss on their generalization accuracy, we pick a reference model depth (red) and use checkpoints (black) from deeper models (blue) which have equal validation perplexity as the reference model does at the end of its pretraining. We then fine-tune these `pretraining-corrected' checkpoints on the compositional tasks. \textbf{(b)} Even when fine-tuning checkpoints with equal validation perplexity, deeper models still generalize better than shallower models do up through six layers. The figure shows generalization accuracies from 134M-parameter models on COGS (single run per condition).}
    \label{fig:pretrain_corr}
\end{figure*}

First, to correct for the deeper models' lower pretraining loss, we repeat our fine-tuning experiments using intermediate checkpoints of pretrained models that have equal validation perplexity within a size class. We pick the least-performant (i.e., shallowest) model within a size class as the reference model and note its validation perplexity at the end of pretraining. We then pick the intermediate checkpoints of all deeper\footnote{We only consider models deeper than the reference model since, in general, shallower models will never attain the perplexity of the reference model at the end of its pretraining. This assumption breaks down when considering the deepest models in each size class, but these are far deeper than the depth at which compositional generalization performance saturates, so we do not extensively explore this regime.} models at the point during pretraining when they achieved this reference perplexity (\Cref{fig:equal_ppl_points}). Finally, we fine-tune each of these checkpoints on the compositional generalization tasks. We repeat this process for successively deeper reference models. We find that even when fine-tuning from checkpoints of equal validation perplexity, deeper models still generalize better than shallower models (\Cref{fig:loss_corr_pt}). 

Next, to correct for the fact that deeper models perform better than shallower ones on the in-distribution split of the compositional generalization tasks, we compare the models' generalization accuracy at points during fine-tuning when they have equal in-distribution loss. \Cref{fig:loss_corr_ft} shows that even after adjusting for in-distribution performance, deeper models still achieve higher accuracy on the out-of-distribution generalization set than shallower models do.

\section{Training and Inference Latency} \label{sec:compute-cost}

\begin{figure}[!ht]
    \centering
    \includegraphics[height=1.4in]{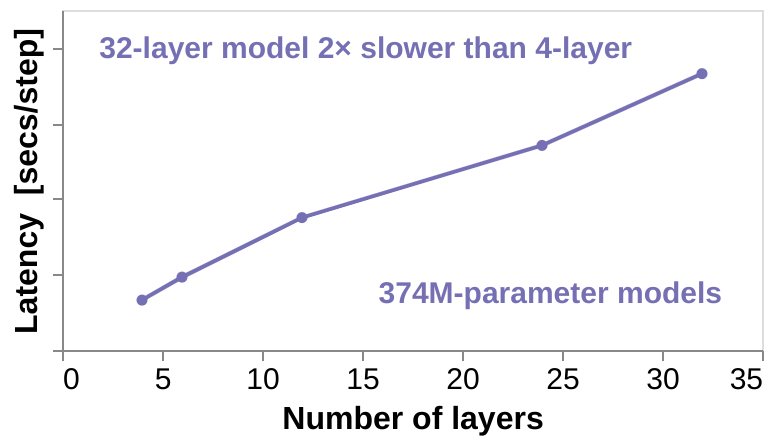}
    \caption{Deeper models train more slowly than shallower ones when controlling for total parameter count. We report relative latency [seconds per training step] for a subset of our 374M-parameter models that were trained on the same accelerator, showing a strongly-linear relationship between depth and latency. Similar relative trends are observed for other model sizes classes.}
    \label{fig:latency-main}
\end{figure}

What are the practical implications of the fact that the benefit of depth saturates after a handful of layers? Empirically, the compute cost of training and running our equal-parameter models exhibits a strongly linear relationship with depth. \Cref{fig:latency-main} shows (for our largest models) that the latency during training grows linearly with the depth of the model. The causes of this penalty are two-fold. First, our choice to use narrower feed-forward dimension for deeper models to maintain constant total parameter count leads to a slightly higher floating point operation (FLOP) count for deeper models. To see this, we start from the cost formula introduced in \citet{kaplan-2020-ScalingModels}:
$$
C_\text{forward} = 2N + 2n_\text{layers}n_\text{ctx}d_\text{attn},
$$
where $N$ is total parameter count. Since both $n_\text{ctx}$ and $d_\text{attn}$ are constant for all models of a particular size (as $d_\text{attn}$ and $d_\text{ff}$ are decoupled from one another), the total FLOP count of a model is linear in depth, though this term is dominated by the $2N$ term unless model depth, attention size, or context length become very large.

The second and likely more important reason that deeper models are slower is because the computations at layer $k$ depend on the results of the computations at layer $k-1$. Because of this sequential dependency, parallelism cannot be applied across layers \cite{Tay2021-qj}.

Combined with the diminishing utility of depth for performance, the linear latency cost of depth as total size is kept constant leads us to the following practical recommendations:
\begin{enumerate}[leftmargin=*]
    \item When trying to minimize GPU-hours for fixed data volume, shallower models can train in far less time than deeper models while still attaining acceptable levels of performance relative to the best-performing model in a given size class.
    \item With a fixed budget of GPU-hours for training, shallower models can train on more data than deeper models can over any fixed amount of time, since shallower models have lower latency, potentially resulting in better performance than deeper models trained on less data. 
\end{enumerate}

These benefits are not confined to training: deeper models also incur a per-layer cost during inference. This means that the penalty a deeper model pays must be amortized over the lifetime cost of using a model for inference. Here too, reducing the GPU-time spent on inference by using shallower models of comparable performance could reduce compute costs, both for cloud-served models and for on-device inference \cite{strubell-etal-2019-energy,pope2022efficiently,gupta-2022-compression}.

\section{Analysis of Feed-Forward Transforms} \label{sec:rank}

At extreme feed-forward ratios, we observe that model performance degrades. We investigate the role that the feed-forward ratio plays in the observed performance of our models. When $\dff$ is smaller than $\dmodel$, this transform is lossy, but small values of $\dff$ could nevertheless impact performance even when $\dff$ is still larger than $\dmodel$. 

\begin{figure*}[ht]
    \centering
    \includegraphics[height=1.6in]{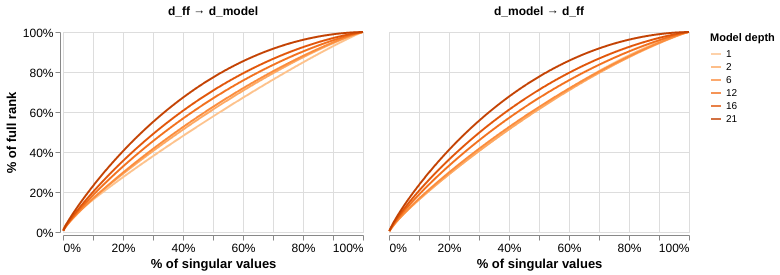}
    \caption{As models get deeper and $\dmodel/\dff$ ratio gets larger (values between $0.01$ for the shallowest model shown and $1$ for the deepest), the input \figleft{} and output \figright{} projections in the feed-forward block become increasingly close to rank-deficient transforms. A graph of $y=x$ here would indicate that models spread their rank equally across all singular values. Data from \SI{134}{\M}-parameter models.}
    \label{fig:cumulative_rank_main}
\end{figure*}

To determine if that is the case, we conduct rank analysis on the affine transforms which comprise the feed-forward block. For a given affine transform $T$, we compute the ordered singular values $\{\sigma_1, \sigma_2, \dots, \sigma_k\}$ where $k = \min(\dmodel, \dff)$ is the rank of $T$ and $\sigma_i \geq \sigma_{i+1}$. We then normalize each value by dividing by the $\ell_1$ norm of $\{\sigma_1, \dots, \sigma_k\}$ to calculate how much of $T$'s image is accounted for by the best $i$-rank approximation of $T$ for $i \leq k$. \Cref{fig:cumulative_rank_main} shows how in deeper models (i.e., those with increasingly large $\dmodel/\dff$ ratios) the transforms become increasingly skewed away from making full-use of their available ranks.

\section{Related Work} \label{sec:related_work}

\paragraph{Compositionality.} Previous work has explored the degree to which neural models exhibit compositional behavior by training or fine-tuning models on compositional tasks such as simple command sequences \citep{lake-2018-GeneralizationNetworks} or semantic parsing \citep{kim-2020-COGSInterpretation, keysers-2020-MeasuringData}. Other work has explored methods to improve the compositional behavior of models, including through data augmentation \citep{qiu-2022-ImprovingAugmentation}, larger models \citep{qiu-2022-EvaluatingParsing}, and architectural changes \citep{gordon-2019-PermutationLanguage,csordas-2021-DevilTransformers-a,ontanon-2022-MakingTasks}. Our work complements these approaches by exploring a specific architecture change: increasing depth without changing total model size.

\paragraph{Impacts of depth.} Theoretical work has shown that the expressive capacity of neural networks in general \citep{raghu-2017-ExpressiveNetworks} and transformer models in particular \citep{merrill-2021-SaturatedCircuits} grows exponentially in depth. Empirical work also points to the role of depth in model performance. In a more general setting, \citet{Tay2021-qj} found that scaling by depth is generally more helpful than scaling by width on downstream tasks, though they do not attempt to control for size. For compositional generalization in particular, \citet{mueller-2022-ColoringModels} found that reducing depth was more harmful than reducing width for pretrained encoder-decoder models. \citet{murty-2022-CharacterizingProjections} observed that deeper transformer encoders often have more tree-like representations and higher parsing accuracies on some compositional tasks. Tempering these positive results, \citet{veit-2016-ResidualNetworks} noted that in models with residual connections, even very deep networks leveraged only shallow subnetworks of roughly constant depth. \citet{brown-2022-Wide} also concluded that wide, shallow transformer models can attain roughly-equal performance to deeper ones. Both sets of results, however, are confounded by a lack of control for total parameter count. Very recently, \citet{gromov-2024-depth} found that nearly half of the layers of deep language models can be pruned after training without substantially harming performance on downstream tasks.

\paragraph{Early-exit schemes.} 
Early-exit research (\citealt{zhou2020bert,schuster2022confident}, \emph{inter alia}) shows that deep models can dynamically be made shallower by skipping later layers of computation once a heuristic deems a computed representation ``good enough.'' We view our work as complementing this approach; while early-exit reduces the computational cost of a model's inference on an input-dependent basis, our work shows that the cost can be reduced for all inputs during both training and inference. However, since we do not explore early-exit training or inference with our equal-parameter models here, it is possible that even our shallower models could benefit from early-exit schemes that would further reduce computational costs. 

\paragraph{Controlling for model size.} There are different possible approaches to studying the impact of hyperparameter choices without affecting the net model size. \citet{kaplan-2020-ScalingModels} covaried number of layers $\nlayers$ with the contextual embedding dimension $\dmodel$, which they coupled to the attention-internal $\dattn$ and feed-forward dimension at the standard ratio of $\dmodel = \dattn = \dff/4$. They concluded that performance increases are largely driven by increasing the total parameter count of models, and that within ``reasonable limits'' language modeling perplexity is only weakly dependent on shape (though \citealt{Tay2021-qj} concluded that the same was not true for performance on downstream tasks, but did so without controlling for the impact of size). Our work investigates the role that depth plays on both pretraining and fine-tuning tasks while controlling for total parameter count.

\section{Conclusion}

Compositional generalization is essential for interpreting novel sentences. What aspects of the transformer LM architecture contribute to an inductive bias favoring compositional generalization? In a controlled experiment that teases apart depth from total number of parameters, we find that deeper transformers show better compositional generalization, and better language modeling performance, independent of their total number of parameters. At the same time, in most cases the usefulness of adding layers decreases rapidly as models get deeper: comparatively shallow models can achieve generalization accuracy on compositional tasks that is comparable to that of much deeper models, and language modeling perplexity within a few percentage points of the best-in-class model. Because deeper transformers have higher latency, this indicates that for a given parameter budget, shallower models can be significantly faster with a minimal sacrifice in performance.

\section{Limitations}

\paragraph{Attention heads.} We do not investigate the role that attention heads play in compositional generalization broadly, nor how the function of heads changes with depth. Previous work (\citealt{michel2019sixteen}, \emph{inter alia}) showed that reducing the number of attention heads in a transformer (pre- or post-training) does not significantly harm performance. Mechanistic interpretability work has found that specific attention heads in transformers learn to compute task-specific functions \cite{voita2019analyzing,htut2019attention,olsson2022incontext}. Our findings here raise two questions which should be further investigated: first, does the relative unimportance of the \emph{number} of attention heads still hold in regimes when a model is significantly shallower and wider than convention; and second, do any of the attention heads in our models learn to perform specifically compositional computations, and does this vary as models get deeper or shallower?

\paragraph{Alternative approaches to controlling for total size.} Our approach to controlling for total parameter count necessitates making depth-width trade-offs. An alternative approach would be to construct Universal Transformers \citep{dehghani-2018-UniversalTransformers}, where each model in a size class has a transformer layer with the same parameters repeated $\nlayers$ times. Such a weight-sharing approach would allow for deeper models to have arbitrarily-wide feed-forward networks, mitigating the impact of making models too narrow. While such weight sharing prevents models from performing different computation in different layers, such restriction may in fact be beneficial for compositional generalization where similar computations (e.g., combining two syntactic phrases to a larger phrase) may need to apply recursively at different scales.

\paragraph{Pretraining corpus effects.} We consider models pretrained on natural-language data. For our particular choice of compositional generalization experiments, the presence of lexical items in both the pretraining corpus and the generalization datasets represents a potential confounder of generalization performance which could be mitigated by modifying compositional datasets \citep{kim-2022-UncontrolledModels}. More generally, we do not study how the distribution of pretraining data affects the inductive biases conferred to LMs \citep{papadimitriou-2023-PretrainLearning}. As a particular area of interest for future work, we point out the hypothesis that including source code in the pretraining corpus \citep{openai-2023-GPT-4Report,google-2023-PaLMReport} will improve compositional generalization.

\paragraph{Fine-tuning vs. in-context learning.} We use fine-tuning to adapt our pretrained models to the compositional tasks. Due to its computational cost and task-specificity, fine-tuning is less useful in practice than in-context learning as model size grows \citep{brown-2020-LanguageLearners}. Because in-context learning only becomes reliable at scales far larger than we are able to train, we did not explore the effect of depth on compositional generalization accuracy in in-context learning \citep{si-2022-PromptingReliable}; we point this out as an avenue for future research. 

\section{Ethics Statement}

Throughout our experimental process, we sought to comply with best practices to mitigate any risks associated with LLM research. We use open-source datasets which are inspectable by third-parties for issues such as bias, and toxicity. We do not release any public checkpoints for the models we train, so there is no risk to misuse of any created artifacts, though we note that we derive our implemented models from existing publicly-available T5 models. We train models on English-only natural-language data, and fuller exploration should be done to explore how language impacts the results found here.

\section*{Acknowledgements} We thank Pete Shaw and Slav Petrov for helpful feedback on previous versions of this paper.

\bibliographystyle{acl_natbib}
\bibliography{anthology,custom}

\appendix
\section{Design and Result Tables} \label{sec:result-tables}

Table~\ref{tab:iso_classes} reports exact hyperparameters for the model classes trained. Table~\ref{tab:compgen-results} displays pretraining and compositional generalization accuracy on all model sizes and tasks.

\begin{table*}[t] \tiny
    \centering
    \resizebox{\textwidth}{!}{
    \begin{tabularx}{\textwidth}{@{}rXXXXXXXXXXXXXXXXXXXXXXXXXX} \toprule
         & \multicolumn{7}{c}{41M} & \multicolumn{10}{c}{134M} & \multicolumn{9}{c}{374M} \\
        \cmidrule(lr){2-8} \cmidrule(lr){9-18} \cmidrule(lr){19-27}
        $\nlayers$ & 1 & \textbf{2} & 3 & 4 & 5 & 6 & 7 & 1 & 2 & 4 & 6 & 8 & \textbf{12} & 16 & 21 & 26 & 32 & 1 & 2 & 4 & 6 & 8 & 12 & 16 & \textbf{24} & 32 \\ 
        $\dff$ & 4779 & \textbf{2048} & 1138 & 682 & 409 & 227 & 97 & 36k & 17k & 8193 & 5121 & 3584 & \textbf{2048} & 1280 & 731 & 393 & 128 & 99k & 49k & 24k & 15k & 11k & 6998 & 4907 & \textbf{2816} & 1770 \\[1ex]
         & \multicolumn{7}{c}{$\dmodel = \dattn =512$,\,\, $\nheads = 8$} & \multicolumn{10}{c}{$\dmodel = \dattn =768$,\,\, $\nheads=8$} & \multicolumn{9}{c}{$\dmodel = \dattn =1024$,\,\, $\nheads=64$} \\[1ex]
         \bottomrule
    \end{tabularx}
    }
    \caption{Models of varying depths across three size classes. Bolded variants are the baseline models whose hyperparameters were taken from \citet{kim-2020-COGSInterpretation} and \citet{raffel-2019-ExploringTransformer}.}
    \label{tab:iso_classes}
\end{table*}

\begin{table*}[ht]
    \centering
    \resizebox{\textwidth}{!}{
    \begin{tabular}{clSSSSS} \toprule
        {\emph{size}} & {$\nlayers$} & {C4 val. PPL ($\downarrow$)} & {COGS ($\uparrow$)} & {COGS-vf ($\uparrow$)} & {GeoQuery Standard ($\uparrow$)} & {English Passivization ($\uparrow$)} \\ \midrule
       \multirow{7}{*}{41M} 
       & 1 & 45.7 & 12.4 & 25.7 & 68.2 & 0.00 \\
       & 2 & 31.1 & 58.2 & 78.3 & 76.4 & 9.88 \\
       & 3 & 29.3 & 63.1 & 80.8 & \bfseries 79.6 & 26.2 \\
       & 4 & 28.8 & 68.5 & 82.5 & 78.6 & 28.0 \\
       & 5 & \bfseries 28.8 & 63.4 & 82.5 & 76.8 & \bfseries 89.9 \\
       & 6 & 29.1 & 68.4 & 82.6 & 77.5 & 74.1 \\
       & 7 & 29.6 & \bfseries 72.3 & \bfseries 83.0 & 77.1 & 78.3 \\ \midrule
       \multirow{10}{*}{134M} 
       & 1 & 33.6 & 19.4 & 26.3 & 72.5 & 0.00 \\
       & 2 & 22.3 & 65.5 & 83.0 & 81.4 & 29.9 \\
       & 4 & 19.4 & 71.1 & 83.6 & 78.2 & 59.3 \\
       & 6 & 18.7 & 74.3 & 83.2 & 80.0 & 49.4 \\
       & 8 & 18.3 & 72.9 & 83.7 & 73.6 & 91.9 \\
       & 12 & \bfseries 18.1 & 73.0 & 84.7 & \bfseries 82.9 & 87.1 \\
       & 16 & 18.2 & 75.0 & 83.8 & 81.1 & 93.2 \\
       & 21 & 18.3 & 75.1 & \bfseries 84.8 & 80.0 & 88.1 \\
       & 26 & 18.6 & 75.4 & 84.1 & 82.1 & \bfseries 98.4 \\ 
       & 32 & 19.2 & \bfseries 75.7 & 84.0 & 78.9 & 94.8 \\ \midrule
       \multirow{9}{*}{374M} 
       & 1 & 28.4 & 21.5 & 36.8 & 72.9 & 0.00 \\
       & 2 & 18.6 & 66.2 & 82.2 & 80.7 & 13.6 \\
       & 4 & 15.9 & 72.4 & 71.9 & 80.0 & 89.8 \\
       & 6 & 15.2 & 75.1 & 83.1 & 78.2 & 18.8 \\
       & 8 & 14.9 & 75.2 & 82.6 & 80.7 & 84.3 \\
       & 12 & 14.6 & 76.3 & 84.3 & 80.0 & 81.0 \\
       & 16 & 14.5 & 76.3 & \bfseries 85.1 & 81.1 & 87.2 \\
       & 24 & \bfseries 14.4 & 78.0 & 83.1 & 83.2 & 89.6 \\ 
       & 32 & 14.7 & \bfseries 78.8 & 79.7 & \bfseries 84.6 & \bfseries 90.2 \\ \bottomrule
    \end{tabular}
    }
    \caption{Validation perplexity ($\downarrow$, lower is better) on C4 after pretraining and generalization accuracy (\%; $\uparrow$, higher is better) on compositional datasets after \SI{10}{k} steps of fine-tuning. Bold values indicate best-in-size-class performance. Data is from a single run per condition.}
    \label{tab:compgen-results}
\end{table*}

\section{Annotated Transformer Layer}\label{sec:layer-diagram}

\begin{figure*}[!htp]
    \centering
    \includegraphics[width=\textwidth]{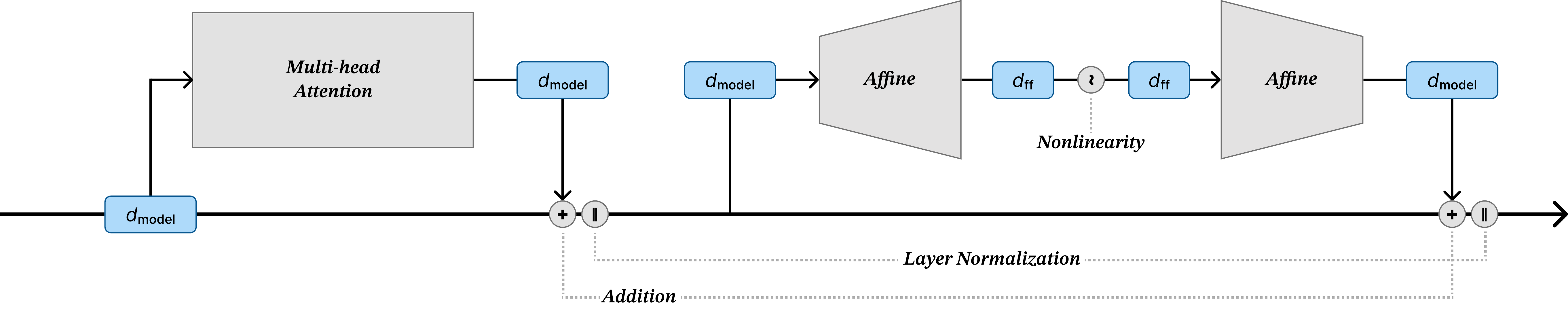}
    \caption{Diagram of a single transformer layer, annotated with the dimensions (blue) of each vector. Information is passed from left to right, through each component (grey box), and added back to the residual embeddings before normalization.}
    \label{fig:layer-diagram}
\end{figure*}

\Cref{fig:layer-diagram} shows the schematic for a single transformer layer. The layers input enters on the left and passes through the various model components (grey boxes), being combined with the residual connections before exiting right to subsequent layers. Blue boxes show the dimensionality of the vectors after transformation; we are primarily concerned with the size of the embedding vectors $\dmodel$ and the internal dimension of the feed-forward block $\dff$. The size of the vectors internal to the attention mechanism, $\dattn$, is not shown here but is usually set to be equal with $\dmodel$; we follow this convention here. Non-learned operations like addition, layer normalization, and the feed-forward network's nonlinearity are shown in grey circles.

%% file: impact-of-depth.bbl
\begin{thebibliography}{44}
\expandafter\ifx\csname natexlab\endcsname\relax\def\natexlab#1{#1}\fi

\bibitem[{Brown et~al.(2022)Brown, Zhao, Shumailov, and
  Mullins}]{brown-2022-Wide}
Jason~Ross Brown, Yiren Zhao, Ilia Shumailov, and Robert~D Mullins. 2022.
\newblock \href {http://arxiv.org/abs/2210.00640} {{Wide attention is the way
  forward for Transformers?}}
\newblock In \emph{Workshop: All Things Attention: Bridging Different
  Perspectives on Attention}.

\bibitem[{Brown et~al.(2020)Brown, Mann, Ryder, Subbiah, Kaplan, Dhariwal,
  Neelakantan, Shyam, Sastry, Askell, Agarwal, Herbert-Voss, Krueger, Henighan,
  Child, Ramesh, Ziegler, Wu, Winter, Hesse, Chen, Sigler, Litwin, Gray, Chess,
  Clark, Berner, McCandlish, Radford, Sutskever, and
  Amodei}]{brown-2020-LanguageLearners}
Tom Brown, Benjamin Mann, Nick Ryder, Melanie Subbiah, Jared~D Kaplan, Prafulla
  Dhariwal, Arvind Neelakantan, Pranav Shyam, Girish Sastry, Amanda Askell,
  Sandhini Agarwal, Ariel Herbert-Voss, Gretchen Krueger, Tom Henighan, Rewon
  Child, Aditya Ramesh, Daniel Ziegler, Jeffrey Wu, Clemens Winter, Chris
  Hesse, Mark Chen, Eric Sigler, Mateusz Litwin, Scott Gray, Benjamin Chess,
  Jack Clark, Christopher Berner, Sam McCandlish, Alec Radford, Ilya Sutskever,
  and Dario Amodei. 2020.
\newblock \href
  {https://papers.nips.cc/paper/2020/hash/1457c0d6bfcb4967418bfb8ac142f64a-Abstract.html}
  {{Language Models are Few-Shot Learners}}.
\newblock In \emph{{Advances in Neural Information Processing Systems 33}},
  volume~33, page 1877–1901. Curran Associates, Inc.

\bibitem[{Csordás et~al.(2021)Csordás, Irie, and
  Schmidhuber}]{csordas-2021-DevilTransformers-a}
Róbert Csordás, Kazuki Irie, and Juergen Schmidhuber. 2021.
\newblock \href {https://doi.org/10.18653/v1/2021.emnlp-main.49} {{The devil is
  in the detail: Simple tricks improve systematic generalization of
  transformers}}.
\newblock In \emph{{Proceedings of the 2021 Conference on Empirical Methods in
  Natural Language Processing}}, pages 619--634, Stroudsburg, PA, USA.
  Association for Computational Linguistics.

\bibitem[{Dehghani et~al.(2018)Dehghani, Gouws, Vinyals, Uszkoreit, and
  Kaiser}]{dehghani-2018-UniversalTransformers}
Mostafa Dehghani, Stephan Gouws, Oriol Vinyals, Jakob Uszkoreit, and Łukasz
  Kaiser. 2018.
\newblock \href {https://arxiv.org/abs/1807.03819} {{Universal Transformers}}.
\newblock In \emph{{International Conference on Learning Representations}}.

\bibitem[{Devlin et~al.(2019)Devlin, Chang, Lee, and
  Toutanova}]{devlin-2019-BERTUnderstanding}
Jacob Devlin, Ming-Wei Chang, Kenton Lee, and Kristina Toutanova. 2019.
\newblock \href {https://doi.org/10.18653/v1/N19-1423} {{BERT: Pre-training of
  Deep Bidirectional Transformers for Language Understanding}}.
\newblock In \emph{{Proceedings of the 2019 Conference of the North American
  Chapter of the Association for Computational Linguistics: Human Language
  Technologies, Volume 1 (Long and Short Papers)}}, page 4171–4186,
  Minneapolis, Minnesota. Association for Computational Linguistics.

\bibitem[{Fodor and Pylyshyn(1988)}]{fodor-1988-ConnectionismAnalysis}
Jerry~A Fodor and Zenon~W Pylyshyn. 1988.
\newblock \href {https://doi.org/10.1016/0010-0277(88)90031-5} {{Connectionism
  and cognitive architecture: a critical analysis}}.
\newblock \emph{Cognition}, 28(1-2):3--71.

\bibitem[{{Google} et~al.(2023){Google}, Anil, Dai, Firat, Johnson, Lepikhin,
  Passos, Shakeri, Taropa, Bailey, Chen, Chu, Clark, Shafey, Huang,
  Meier-Hellstern, Mishra, Moreira, Omernick, Robinson, Ruder, Tay, Xiao, Xu,
  Zhang, Abrego, Ahn, Austin, Barham, Botha, Bradbury, Brahma, Brooks, Catasta,
  Cheng, Cherry, Choquette-Choo, Chowdhery, Crepy, Dave, Dehghani, Dev, Devlin,
  Díaz, Du, Dyer, Feinberg, Feng, Fienber, Freitag, Garcia, Gehrmann,
  Gonzalez, Gur-Ari, Hand, Hashemi, Hou, Howland, Hu, Hui, Hurwitz, Isard,
  Ittycheriah, Jagielski, Jia, Kenealy, Krikun, Kudugunta, Lan, Lee, Lee, Li,
  {Li, Music}, Li, Li, Li, Lim, Lin, Liu, Liu, Maggioni, Mahendru, Maynez,
  Misra, Moussalem, Nado, Nham, Ni, Nystrom, Parrish, Pellat, Polacek, Polozov,
  Pope, Qiao, Reif, Richter, Riley, Ros, Roy, Saeta, Samuel, Shelby, Slone,
  Smilkov, So, Sohn, Tokumine, Valter, Vasudevan, Vodrahalli, Wang, Wang, Wang,
  Wang, Wieting, Wu, Xu, Xu, Xue, Yin, Yu, Zhang, Zheng, Zheng, Zhou, Zhou,
  Petrov, and Wu}]{google-2023-PaLMReport}
{Google}, Rohan Anil, Andrew~M Dai, Orhan Firat, Melvin Johnson, Dmitry
  Lepikhin, Alexandre Passos, Siamak Shakeri, Emanuel Taropa, Paige Bailey,
  Zhifeng Chen, Eric Chu, Jonathan~H Clark, Laurent~El Shafey, Yanping Huang,
  Kathy Meier-Hellstern, Gaurav Mishra, Erica Moreira, Mark Omernick, Kevin
  Robinson, Sebastian Ruder, Yi~Tay, Kefan Xiao, Yuanzhong Xu, Yujing Zhang,
  Gustavo~Hernandez Abrego, Junwhan Ahn, Jacob Austin, Paul Barham, Jan Botha,
  James Bradbury, Siddhartha Brahma, Kevin Brooks, Michele Catasta, Yong Cheng,
  Colin Cherry, Christopher~A Choquette-Choo, Aakanksha Chowdhery, Clément
  Crepy, Shachi Dave, Mostafa Dehghani, Sunipa Dev, Jacob Devlin, Mark Díaz,
  Nan Du, Ethan Dyer, Vlad Feinberg, Fangxiaoyu Feng, Vlad Fienber, Markus
  Freitag, Xavier Garcia, Sebastian Gehrmann, Lucas Gonzalez, Guy Gur-Ari,
  Steven Hand, Hadi Hashemi, Le~Hou, Joshua Howland, Andrea Hu, Jeffrey Hui,
  Jeremy Hurwitz, Michael Isard, Abe Ittycheriah, Matthew Jagielski, Wenhao
  Jia, Kathleen Kenealy, Maxim Krikun, Sneha Kudugunta, Chang Lan, Katherine
  Lee, Benjamin Lee, Eric Li, {Li, Music}, Wei Li, Yaguang Li, Jian Li,
  Hyeontaek Lim, Hanzhao Lin, Zhongtao Liu, Frederick Liu, Marcello Maggioni,
  Aroma Mahendru, Joshua Maynez, Vedant Misra, Maysam Moussalem, Zachary Nado,
  John Nham, Eric Ni, Andrew Nystrom, Alicia Parrish, Marie Pellat, Martin
  Polacek, Alex Polozov, Reiner Pope, Siyuan Qiao, Emily Reif, Bryan Richter,
  Parker Riley, Alex~Castro Ros, Aurko Roy, Brennan Saeta, Rajkumar Samuel,
  Renee Shelby, Ambrose Slone, Daniel Smilkov, David~R So, Daniel Sohn, Simon
  Tokumine, Dasha Valter, Vijay Vasudevan, Kiran Vodrahalli, Xuezhi Wang,
  Pidong Wang, Zirui Wang, Tao Wang, John Wieting, Yuhuai Wu, Kelvin Xu, Yunhan
  Xu, Linting Xue, Pengcheng Yin, Jiahui Yu, Qiao Zhang, Steven Zheng,
  Ce~Zheng, Weikang Zhou, Denny Zhou, Slav Petrov, and Yonghui Wu. 2023.
\newblock \href {https://arxiv.org/abs/2305.10403} {{PaLM 2 Technical Report}}.
\newblock Technical report, Google.

\bibitem[{Gordon et~al.(2019)Gordon, Lopez-Paz, Baroni, and
  Bouchacourt}]{gordon-2019-PermutationLanguage}
Jonathan Gordon, David Lopez-Paz, Marco Baroni, and Diane Bouchacourt. 2019.
\newblock \href {https://openreview.net/forum?id=SylVNerFvr} {{Permutation
  Equivariant Models for Compositional Generalization in Language}}.
\newblock In \emph{{ICLR 2020 (OpenReview)}}.

\bibitem[{Gromov et~al.(2024)Gromov, Tirumala, Shapourian, Glorioso, and
  Roberts}]{gromov-2024-depth}
Andrey Gromov, Kushal Tirumala, Hassan Shapourian, Paolo Glorioso, and
  Daniel~A. Roberts. 2024.
\newblock \href {https://arxiv.org/abs/2403.17887v1} {The unreasonable
  ineffectiveness of the deeper layers}.
\newblock Technical Report MIT-CTP/5694, Center for Theoretical Physics,
  Massachusetts Institute of Technology, Cambridge, MA.

\bibitem[{Gupta and Agrawal(2022)}]{gupta-2022-compression}
Manish Gupta and Puneet Agrawal. 2022.
\newblock \href {https://doi.org/10.1145/3487045} {Compression of deep learning
  models for text: A survey}.
\newblock \emph{ACM Trans. Knowl. Discov. Data}, 16(4).

\bibitem[{Hoffmann et~al.(2022)Hoffmann, Borgeaud, Mensch, Buchatskaya, Cai,
  Rutherford, Casas, Hendricks, Welbl, Clark, Hennigan, Noland, Millican,
  van~den Driessche, Damoc, Guy, Osindero, Simonyan, Elsen, Rae, Vinyals, and
  Sifre}]{hoffmann-2022-TrainingModels}
Jordan Hoffmann, Sebastian Borgeaud, Arthur Mensch, Elena Buchatskaya, Trevor
  Cai, Eliza Rutherford, Diego de~Las Casas, Lisa~Anne Hendricks, Johannes
  Welbl, Aidan Clark, Tom Hennigan, Eric Noland, Katie Millican, George van~den
  Driessche, Bogdan Damoc, Aurelia Guy, Simon Osindero, Karen Simonyan, Erich
  Elsen, Jack~W Rae, Oriol Vinyals, and Laurent Sifre. 2022.
\newblock \href {https://doi.org/10.48550/arXiv.2203.15556} {{Training
  Compute-Optimal Large Language Models}}.

\bibitem[{Htut et~al.(2019)Htut, Phang, Bordia, and Bowman}]{htut2019attention}
Phu~Mon Htut, Jason Phang, Shikha Bordia, and Samuel~R. Bowman. 2019.
\newblock \href {http://arxiv.org/abs/1911.12246} {Do attention heads in bert
  track syntactic dependencies?}

\bibitem[{Kaplan et~al.(2020)Kaplan, McCandlish, Henighan, Brown, Chess, Child,
  Gray, Radford, Wu, and Amodei}]{kaplan-2020-ScalingModels}
Jared Kaplan, Sam McCandlish, Tom Henighan, Tom~B Brown, Benjamin Chess, Rewon
  Child, Scott Gray, Alec Radford, Jeffrey Wu, and Dario Amodei. 2020.
\newblock \href {http://arxiv.org/abs/2001.08361} {{Scaling laws for neural
  language models}}.

\bibitem[{Keysers et~al.(2020)Keysers, Schärli, Scales, Buisman, Furrer,
  Kashubin, Momchev, Sinopalnikov, Stafiniak, Tihon, Tsarkov, Wang, van Zee,
  and Bousquet}]{keysers-2020-MeasuringData}
Daniel Keysers, Nathanael Schärli, Nathan Scales, Hylke Buisman, Daniel
  Furrer, Sergii Kashubin, Nikola Momchev, Danila Sinopalnikov, Lukasz
  Stafiniak, Tibor Tihon, Dmitry Tsarkov, Xiao Wang, Marc van Zee, and Olivier
  Bousquet. 2020.
\newblock \href {https://doi.org/10.48550/arXiv.1912.09713} {{Measuring
  Compositional Generalization: A Comprehensive Method on Realistic Data}}.

\bibitem[{Kim and Linzen(2020)}]{kim-2020-COGSInterpretation}
Najoung Kim and Tal Linzen. 2020.
\newblock \href {https://doi.org/10.18653/v1/2020.emnlp-main.731} {{COGS: A
  Compositional Generalization Challenge Based on Semantic Interpretation}}.
\newblock In \emph{{Proceedings of the 2020 Conference on Empirical Methods in
  Natural Language Processing (EMNLP)}}, page 9087–9105, Online. Association
  for Computational Linguistics.

\bibitem[{Kim et~al.(2022)Kim, Linzen, and
  Smolensky}]{kim-2022-UncontrolledModels}
Najoung Kim, Tal Linzen, and Paul Smolensky. 2022.
\newblock \href {http://arxiv.org/abs/2212.10769} {{Uncontrolled lexical
  exposure leads to overestimation of compositional generalization in
  pretrained models}}.

\bibitem[{Lake and Baroni(2018)}]{lake-2018-GeneralizationNetworks}
Brenden Lake and Marco Baroni. 2018.
\newblock \href {https://proceedings.mlr.press/v80/lake18a.html}
  {{Generalization without Systematicity: On the Compositional Skills of
  Sequence-to-Sequence Recurrent Networks}}.
\newblock In \emph{{Proceedings of the 35th International Conference on Machine
  Learning}}, pages 2873--2882. PMLR.

\bibitem[{Merrill et~al.(2021)Merrill, Sabharwal, and
  Smith}]{merrill-2021-SaturatedCircuits}
William Merrill, Ashish Sabharwal, and Noah~A Smith. 2021.
\newblock \href {https://doi.org/10.1162/tacl\_a\_00493} {{Saturated
  Transformers are Constant-Depth Threshold Circuits}}.
\newblock \emph{Transactions of the Association for Computational Linguistics},
  pages 843--856.

\bibitem[{Michel et~al.(2019)Michel, Levy, and Neubig}]{michel2019sixteen}
Paul Michel, Omer Levy, and Graham Neubig. 2019.
\newblock \href
  {https://proceedings.neurips.cc/paper_files/paper/2019/file/2c601ad9d2ff9bc8b282670cdd54f69f-Paper.pdf}
  {Are sixteen heads really better than one?}
\newblock In \emph{Advances in Neural Information Processing Systems},
  volume~32. Curran Associates, Inc.

\bibitem[{Montague(1970)}]{montague-1970-UniversalGrammar}
Richard Montague. 1970.
\newblock \href {https://doi.org/10.1111/j.1755-2567.1970.tb00434.x}
  {{Universal grammar}}.
\newblock \emph{Theoria}, 36(3):373--398.

\bibitem[{Mueller et~al.(2022)Mueller, Frank, Linzen, Wang, and
  Schuster}]{mueller-2022-ColoringModels}
Aaron Mueller, Robert Frank, Tal Linzen, Luheng Wang, and Sebastian Schuster.
  2022.
\newblock \href {https://doi.org/10.18653/v1/2022.findings-acl.106} {{Coloring
  the Blank Slate: Pre-training Imparts a Hierarchical Inductive Bias to
  Sequence-to-sequence Models}}.
\newblock In \emph{{Findings of the Association for Computational Linguistics:
  ACL 2022}}, page 1352–1368, Dublin, Ireland. Association for Computational
  Linguistics.

\bibitem[{Muennighoff et~al.(2023)Muennighoff, Rush, Barak, Scao, Piktus, Tazi,
  Pyysalo, Wolf, and Raffel}]{muennighoff-2023-ScalingModels}
Niklas Muennighoff, Alexander~M Rush, Boaz Barak, Teven~Le Scao, Aleksandra
  Piktus, Nouamane Tazi, Sampo Pyysalo, Thomas Wolf, and Colin Raffel. 2023.
\newblock \href {http://arxiv.org/abs/2305.16264} {{Scaling data-constrained
  language models}}.
\newblock In \emph{37th Conference on Neural Information Processing Systems
  (NeurIPS 2023)}.

\bibitem[{Mulligan et~al.(2021)Mulligan, Frank, and
  Linzen}]{mulligan-2021-StructureTransformations-a}
Karl Mulligan, Robert Frank, and Tal Linzen. 2021.
\newblock \href {https://aclanthology.org/2021.scil-1.12.pdf} {{Structure Here,
  Bias There: Hierarchical Generalization by Jointly Learning Syntactic
  Transformations}}.
\newblock In \emph{{Proceedings of the Society for Computation in Linguistics
  2021}}, pages 125--135.

\bibitem[{Murty et~al.(2023)Murty, Sharma, Andreas, and
  Manning}]{murty-2022-CharacterizingProjections}
Shikhar Murty, Pratyusha Sharma, Jacob Andreas, and Christopher~D Manning.
  2023.
\newblock \href {https://doi.org/10.48550/arXiv.2211.01288} {{Characterizing
  Intrinsic Compositionality in Transformers with Tree Projections}}.
\newblock In \emph{The Eleventh International Conference on Learning
  Representations}.

\bibitem[{Olsson et~al.(2022)Olsson, Elhage, Nanda, Joseph, DasSarma, Henighan,
  Mann, Askell, Bai, Chen, Conerly, Drain, Ganguli, Hatfield-Dodds, Hernandez,
  Johnston, Jones, Kernion, Lovitt, Ndousse, Amodei, Brown, Clark, Kaplan,
  McCandlish, and Olah}]{olsson2022incontext}
Catherine Olsson, Nelson Elhage, Neel Nanda, Nicholas Joseph, Nova DasSarma,
  Tom Henighan, Ben Mann, Amanda Askell, Yuntao Bai, Anna Chen, Tom Conerly,
  Dawn Drain, Deep Ganguli, Zac Hatfield-Dodds, Danny Hernandez, Scott
  Johnston, Andy Jones, Jackson Kernion, Liane Lovitt, Kamal Ndousse, Dario
  Amodei, Tom Brown, Jack Clark, Jared Kaplan, Sam McCandlish, and Chris Olah.
  2022.
\newblock \href {http://arxiv.org/abs/2209.11895} {In-context learning and
  induction heads}.

\bibitem[{Ontanon et~al.(2022)Ontanon, Ainslie, Fisher, and
  Cvicek}]{ontanon-2022-MakingTasks}
Santiago Ontanon, Joshua Ainslie, Zachary Fisher, and Vaclav Cvicek. 2022.
\newblock \href {https://doi.org/10.18653/v1/2022.acl-long.251} {{Making
  transformers solve compositional tasks}}.
\newblock In \emph{{Proceedings of the 60th Annual Meeting of the Association
  for Computational Linguistics (Volume 1: Long Papers)}}, pages 3591--3607,
  Stroudsburg, PA, USA. Association for Computational Linguistics.

\bibitem[{{OpenAI}(2023)}]{openai-2023-GPT-4Report}
{OpenAI}. 2023.
\newblock \href {http://arxiv.org/abs/2303.08774} {{GPT-4 Technical Report}}.

\bibitem[{Papadimitriou and
  Jurafsky(2023)}]{papadimitriou-2023-PretrainLearning}
Isabel Papadimitriou and Dan Jurafsky. 2023.
\newblock \href {https://doi.org/10.18653/v1/2023.findings-emnlp.563}
  {Injecting structural hints: Using language models to study inductive biases
  in language learning}.
\newblock In \emph{Findings of the Association for Computational Linguistics:
  EMNLP 2023}, pages 8402--8413, Singapore. Association for Computational
  Linguistics.

\bibitem[{Pope et~al.(2022)Pope, Douglas, Chowdhery, Devlin, Bradbury,
  Levskaya, Heek, Xiao, Agrawal, and Dean}]{pope2022efficiently}
Reiner Pope, Sholto Douglas, Aakanksha Chowdhery, Jacob Devlin, James Bradbury,
  Anselm Levskaya, Jonathan Heek, Kefan Xiao, Shivani Agrawal, and Jeff Dean.
  2022.
\newblock \href {http://arxiv.org/abs/2211.05102} {Efficiently scaling
  transformer inference}.

\bibitem[{Qiu et~al.(2022{\natexlab{a}})Qiu, Shaw, Pasupat, Nowak, Linzen, Sha,
  and Toutanova}]{qiu-2022-ImprovingAugmentation}
Linlu Qiu, Peter Shaw, Panupong Pasupat, Pawel Nowak, Tal Linzen, Fei Sha, and
  Kristina Toutanova. 2022{\natexlab{a}}.
\newblock \href {https://doi.org/10.18653/v1/2022.naacl-main.323} {Improving
  compositional generalization with latent structure and data augmentation}.
\newblock In \emph{Proceedings of the 2022 Conference of the North American
  Chapter of the Association for Computational Linguistics: Human Language
  Technologies}, pages 4341--4362, Seattle, United States. Association for
  Computational Linguistics.

\bibitem[{Qiu et~al.(2022{\natexlab{b}})Qiu, Shaw, Pasupat, Shi, Herzig,
  Pitler, Sha, and Toutanova}]{qiu-2022-EvaluatingParsing}
Linlu Qiu, Peter Shaw, Panupong Pasupat, Tianze Shi, Jonathan Herzig, Emily
  Pitler, Fei Sha, and Kristina Toutanova. 2022{\natexlab{b}}.
\newblock \href {https://doi.org/10.18653/v1/2022.emnlp-main.624} {Evaluating
  the impact of model scale for compositional generalization in semantic
  parsing}.
\newblock In \emph{Proceedings of the 2022 Conference on Empirical Methods in
  Natural Language Processing}, pages 9157--9179, Abu Dhabi, United Arab
  Emirates. Association for Computational Linguistics.

\bibitem[{Raffel et~al.(2019)Raffel, Shazeer, Roberts, Lee, Narang, Matena,
  Zhou, Li, and Liu}]{raffel-2019-ExploringTransformer}
Colin Raffel, Noam Shazeer, Adam Roberts, Katherine Lee, Sharan Narang, Michael
  Matena, Yanqi Zhou, Wei Li, and Peter~J Liu. 2019.
\newblock \href {http://arxiv.org/abs/1910.10683} {{Exploring the limits of
  transfer learning with a unified text-to-text transformer}}.
\newblock \emph{Journal of machine learning research: JMLR}, 21(2020):1--67.

\bibitem[{Raghu et~al.(2017)Raghu, Poole, Kleinberg, Ganguli, and
  Sohl-Dickstein}]{raghu-2017-ExpressiveNetworks}
Maithra Raghu, Ben Poole, Jon Kleinberg, Surya Ganguli, and Jascha
  Sohl-Dickstein. 2017.
\newblock \href {https://proceedings.mlr.press/v70/raghu17a.html} {{On the
  Expressive Power of Deep Neural Networks}}.
\newblock In \emph{{International Conference on Machine Learning}}, pages
  2847--2854. PMLR.

\bibitem[{Roberts et~al.(2022)Roberts, Chung, Levskaya, Mishra, Bradbury,
  Andor, Narang, Lester, Gaffney, Mohiuddin, Hawthorne, Lewkowycz, Salcianu,
  van Zee, Austin, Goodman, Soares, Hu, Tsvyashchenko, Chowdhery, Bastings,
  Bulian, Garcia, Ni, Chen, Kenealy, Clark, Lee, Garrette, Lee-Thorp, Raffel,
  Shazeer, Ritter, Bosma, Passos, Maitin-Shepard, Fiedel, Omernick, Saeta,
  Sepassi, Spiridonov, Newlan, and Gesmundo}]{roberts2022scaling}
Adam Roberts, Hyung~Won Chung, Anselm Levskaya, Gaurav Mishra, James Bradbury,
  Daniel Andor, Sharan Narang, Brian Lester, Colin Gaffney, Afroz Mohiuddin,
  Curtis Hawthorne, Aitor Lewkowycz, Alex Salcianu, Marc van Zee, Jacob Austin,
  Sebastian Goodman, Livio~Baldini Soares, Haitang Hu, Sasha Tsvyashchenko,
  Aakanksha Chowdhery, Jasmijn Bastings, Jannis Bulian, Xavier Garcia, Jianmo
  Ni, Andrew Chen, Kathleen Kenealy, Jonathan~H. Clark, Stephan Lee, Dan
  Garrette, James Lee-Thorp, Colin Raffel, Noam Shazeer, Marvin Ritter, Maarten
  Bosma, Alexandre Passos, Jeremy Maitin-Shepard, Noah Fiedel, Mark Omernick,
  Brennan Saeta, Ryan Sepassi, Alexander Spiridonov, Joshua Newlan, and Andrea
  Gesmundo. 2022.
\newblock \href {http://arxiv.org/abs/2203.17189} {Scaling up models and data
  with $\texttt{t5x}$ and $\texttt{seqio}$}.

\bibitem[{Schuster et~al.(2022)Schuster, Fisch, Gupta, Dehghani, Bahri, Tran,
  Tay, and Metzler}]{schuster2022confident}
Tal Schuster, Adam Fisch, Jai Gupta, Mostafa Dehghani, Dara Bahri, Vinh Tran,
  Yi~Tay, and Donald Metzler. 2022.
\newblock \href
  {https://proceedings.neurips.cc/paper_files/paper/2022/file/6fac9e316a4ae75ea244ddcef1982c71-Paper-Conference.pdf}
  {Confident adaptive language modeling}.
\newblock In \emph{Advances in Neural Information Processing Systems},
  volume~35, pages 17456--17472. Curran Associates, Inc.

\bibitem[{Si et~al.(2022)Si, Gan, Yang, Wang, Wang, Boyd-Graber, and
  Wang}]{si-2022-PromptingReliable}
Chenglei Si, Zhe Gan, Zhengyuan Yang, Shuohang Wang, Jianfeng Wang, Jordan
  Boyd-Graber, and Lijuan Wang. 2022.
\newblock \href {https://arxiv.org/abs/2210.09150} {{Prompting GPT-3 to be
  reliable}}.
\newblock In \emph{{The Eleventh International Conference on Learning
  Representations}}.

\bibitem[{Strubell et~al.(2019)Strubell, Ganesh, and
  McCallum}]{strubell-etal-2019-energy}
Emma Strubell, Ananya Ganesh, and Andrew McCallum. 2019.
\newblock \href {https://doi.org/10.18653/v1/P19-1355} {Energy and policy
  considerations for deep learning in {NLP}}.
\newblock In \emph{Proceedings of the 57th Annual Meeting of the Association
  for Computational Linguistics}, pages 3645--3650, Florence, Italy.
  Association for Computational Linguistics.

\bibitem[{Tay et~al.(2021)Tay, Dehghani, Rao, Fedus, Abnar, Chung, Narang,
  Yogatama, Vaswani, and Metzler}]{Tay2021-qj}
Yi~Tay, Mostafa Dehghani, Jinfeng Rao, William Fedus, Samira Abnar, Hyung~Won
  Chung, Sharan Narang, Dani Yogatama, Ashish Vaswani, and Donald Metzler.
  2021.
\newblock \href {https://arxiv.org/abs/2109.10686} {{Scale efficiently:
  Insights from pre-training and fine-tuning Transformers}}.
\newblock In \emph{{International Conference on Learning Representations}}.

\bibitem[{Vaswani et~al.(2017)Vaswani, Shazeer, Parmar, Uszkoreit, Jones,
  Gomez, Kaiser, and Polosukhin}]{vaswani-2017-AttentionNeed}
Ashish Vaswani, Noam Shazeer, Niki Parmar, Jakob Uszkoreit, Llion Jones,
  Aidan~N Gomez, Łukasz Kaiser, and Illia Polosukhin. 2017.
\newblock \href {https://arxiv.org/abs/1706.03762} {{Attention is all you
  need}}.
\newblock In \emph{{Proceedings of the 31st International Conference on Neural
  Information Processing Systems}}, NIPS'17, page 6000–6010, Red Hook, NY,
  USA. Curran Associates Inc.

\bibitem[{Veit et~al.(2016)Veit, Wilber, and
  Belongie}]{veit-2016-ResidualNetworks}
Andreas Veit, Michael Wilber, and Serge Belongie. 2016.
\newblock \href {https://arxiv.org/abs/1605.06431} {Residual networks behave
  like ensembles of relatively shallow networks}.
\newblock In \emph{Proceedings of the 30th International Conference on Neural
  Information Processing Systems}, NIPS'16, page 550–558, Red Hook, NY, USA.
  Curran Associates Inc.

\bibitem[{Voita et~al.(2019)Voita, Talbot, Moiseev, Sennrich, and
  Titov}]{voita2019analyzing}
Elena Voita, David Talbot, Fedor Moiseev, Rico Sennrich, and Ivan Titov. 2019.
\newblock \href {https://doi.org/10.18653/v1/P19-1580} {Analyzing multi-head
  self-attention: Specialized heads do the heavy lifting, the rest can be
  pruned}.
\newblock In \emph{Proceedings of the 57th Annual Meeting of the Association
  for Computational Linguistics}, pages 5797--5808, Florence, Italy.
  Association for Computational Linguistics.

\bibitem[{Wang et~al.(2022)Wang, Roberts, Hesslow, Scao, Chung, Beltagy,
  Launay, and Raffel}]{wang-2022-What}
Thomas Wang, Adam Roberts, Daniel Hesslow, Teven~Le Scao, Hyung~Won Chung,
  Iz~Beltagy, Julien Launay, and Colin Raffel. 2022.
\newblock \href {https://proceedings.mlr.press/v162/wang22u.html} {What
  language model architecture and pretraining objective works best for
  zero-shot generalization?}
\newblock In \emph{Proceedings of the 39th International Conference on Machine
  Learning}, volume 162 of \emph{Proceedings of Machine Learning Research},
  pages 22964--22984. PMLR.

\bibitem[{Zelle and Mooney(1996)}]{zelle-1996-LearningProgramming}
John~M Zelle and Raymond~J Mooney. 1996.
\newblock \href {https://doi.org/10.5555/1864519.1864543} {{Learning to parse
  database queries using inductive logic programming}}.
\newblock In \emph{{Proceedings of the thirteenth national conference on
  Artificial intelligence - Volume 2}}, AAAI'96, pages 1050--1055. AAAI Press.

\bibitem[{Zhou et~al.(2020)Zhou, Xu, Ge, McAuley, Xu, and Wei}]{zhou2020bert}
Wangchunshu Zhou, Canwen Xu, Tao Ge, Julian McAuley, Ke~Xu, and Furu Wei. 2020.
\newblock \href
  {https://proceedings.neurips.cc/paper_files/paper/2020/file/d4dd111a4fd973394238aca5c05bebe3-Paper.pdf}
  {{BERT} loses patience: Fast and robust inference with early exit}.
\newblock In \emph{Advances in Neural Information Processing Systems},
  volume~33, pages 18330--18341. Curran Associates, Inc.

\end{thebibliography}
